\documentclass{article} %
\usepackage{colm2024_conference}
\usepackage{booktabs}
\usepackage{siunitx}
\usepackage{graphicx}
\usepackage{enumitem}
\usepackage{wrapfig}
\usepackage{algorithm}
\usepackage{algpseudocode}
\usepackage[T1]{fontenc}
\usepackage{lmodern}
\usepackage{microtype}
\usepackage{amsmath}
\usepackage{colortbl}
\usepackage[utf8]{inputenc}
\usepackage{caption}
\usepackage{subcaption}
\usepackage{setspace}
\usepackage{url}
\usepackage{multirow}
\usepackage{colortbl}
\usepackage{tabularx}
\usepackage{blindtext}
\usepackage{pgfplots}
\pgfplotsset{compat=1.18} 
\usepackage{tikz}
\usetikzlibrary{er,positioning,bayesnet}
\usepackage{makecell}
\usepackage{tipa}
\usepackage{nicefrac}
\usepackage{tocloft}
\usepackage{listings}
\usepackage[raster,skins,breakable]{tcolorbox}
\usepackage{xltabular}
\usepackage{adjustbox}
\usepackage{xurl}
\usepackage[normalem]{ulem}
\usepackage{multirow}
\usepackage{array}
\usepackage{booktabs}
\usepackage{makecell}

\usepackage{amsmath,amsfonts,bm}

\def\eqref#1{(\ref{#1})}

\def\1{\bm{1}}

\DeclareMathAlphabet{\mathsfit}{\encodingdefault}{\sfdefault}{m}{sl}
\SetMathAlphabet{\mathsfit}{bold}{\encodingdefault}{\sfdefault}{bx}{n}

\def \y         {\mathbf{y}}
\def \x         {\mathbf{x}}

\def \ce        {\textrm{CE}}
\def \sft       {\textrm{SFT}}
\def \ppo       {\textrm{PPO}}
\def \old       {\text{old}}
\def \clip      {\text{clip}}
\def \grpo      {\textrm{GRPO}}
\def \mean      {\textrm{mean}}
\def \std       {\textrm{std}}
\def \batch     {\textrm{batch}}
\def \step      {\textrm{step}}

\def \sgrpo     {\textrm{S-GRPO}}
\def \srloo     {\textrm{S-RLOO}}
\def \srf       {\textrm{S-RF++}}

\def\A{{\mathcal{A}}}
\def\S{{\mathcal{S}}}

\def\P{{\mathcal{P}}}
\def\R{{\mathcal{R}}}
\def\M{{\mathcal{M}}}
\def\L{{\mathcal{L}}}
\def\J{{\mathcal{J}}}

\newcommand{\E}{\mathbb{E}}
\newcommand{\D}{\mathcal{D}}

\usepackage{listings}
\usepackage{xcolor}

\definecolor{lightgray}{rgb}{0.9,0.9,0.9}
\definecolor{tableblue}{RGB}{237, 242, 249}
\definecolor{red}{rgb}{0.8,0.0,0.0}  
\definecolor{green}{rgb}{0.0, 0.5, 0.0}

\title{Building Autonomous GUI Navigation via \\ Agentic-Q Estimation and Step-Wise Policy Optimization}

\author{
  \textbf{Yibo Wang}\textsuperscript{\rm 1,2,}\thanks{Work done during the internship at Alibaba Group.} ,
  \textbf{Guangda Huzhang},
  \textbf{Yuwei Hu}\textsuperscript{\rm 2},
  \textbf{Yu Xia}\textsuperscript{\rm 2},
  \textbf{Shiyin Lu}\textsuperscript{\rm 2},\\
  \vspace{2pt}
  \textbf{Qing-Guo Chen}\textsuperscript{\rm 2},
  \textbf{Zhao Xu}\textsuperscript{\rm 2},
  \textbf{Weihua Luo}\textsuperscript{\rm 2},
  \textbf{Kaifu Zhang}\textsuperscript{\rm 2},
  \textbf{Lijun Zhang}\textsuperscript{\rm 1}\\
  \vspace{10pt}
  \textsuperscript{\rm 1}National Key Laboratory for Novel Software Technology, Nanjing University\\
  \vspace{3pt}
  \textsuperscript{\rm 2}Ovis Team, Alibaba Group
}

\begin{document}

\maketitle

\begin{abstract}
  Recent advances in Multimodal Large Language Models (MLLMs) have substantially driven the progress of autonomous agents for Graphical User Interface (GUI). Nevertheless, in real-world applications, GUI agents are often faced with non-stationary environments, leading to high computational costs for data curation and policy optimization. In this report, we introduce a novel MLLM-centered framework for GUI agents, which consists of two components: agentic-Q estimation and step-wise policy optimization. The former one aims to optimize a Q-model that can generate step-wise values to evaluate the contribution of a given action to task completion. The latter one takes step-wise samples from the state-action trajectory as inputs, and optimizes the policy via reinforcement learning with our agentic-Q model. It should be noticed that (i) all state-action trajectories are produced by the policy itself, so that the data collection costs are manageable; (ii) the policy update is decoupled from the environment, ensuring stable and efficient optimization. Empirical evaluations show that our framework endows Ovis2.5-9B with powerful GUI interaction capabilities, achieving remarkable performances on GUI navigation and grounding benchmarks and even surpassing contenders with larger scales.
\end{abstract}

\begin{figure}[H]
  \centering
  \includegraphics[width=0.9\linewidth]{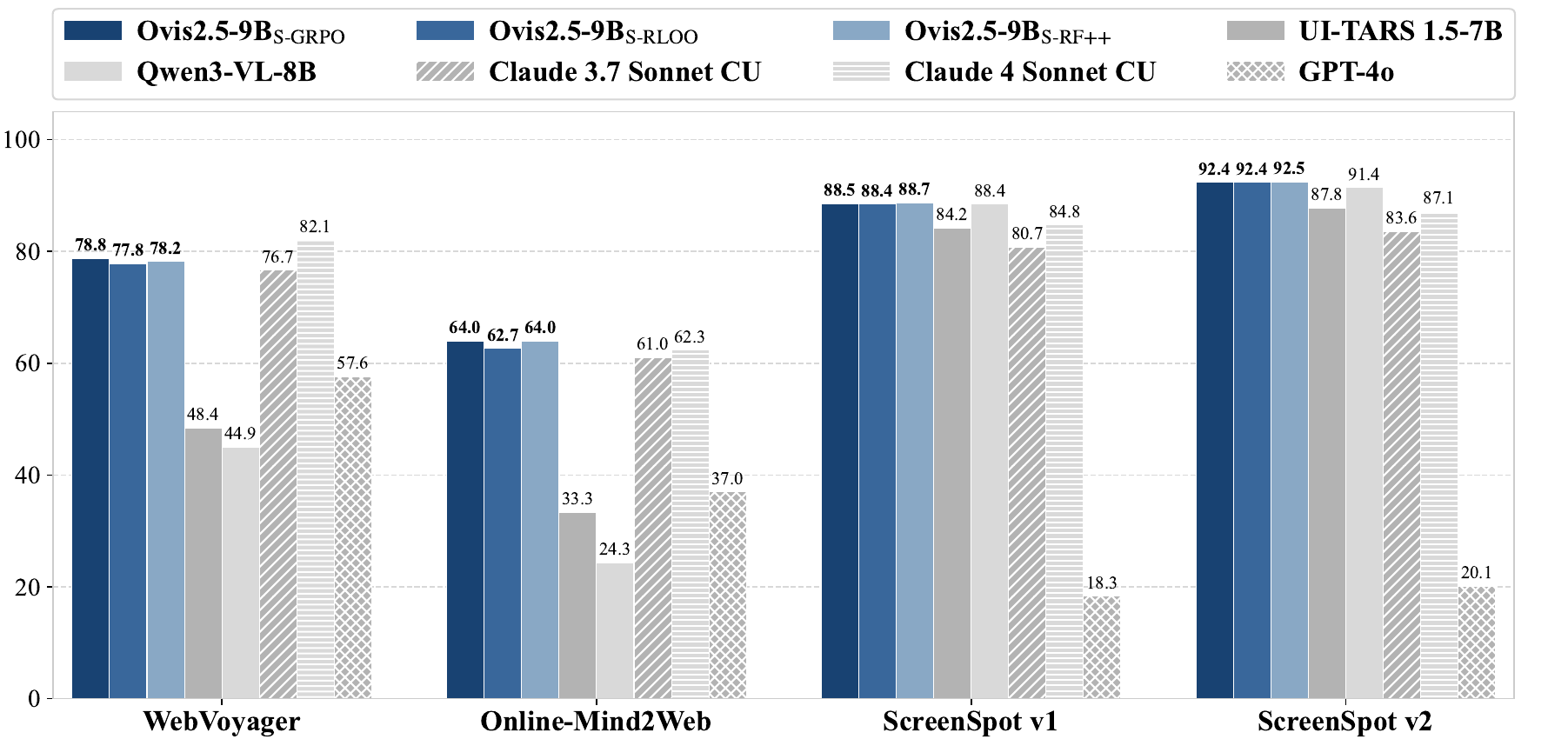}
  \caption{Performance comparisons of agentic Ovis2.5-9B series and other contenders.}
  \label{fig:benchmark}
\end{figure}

\section{Introduction}
\label{sec:intro}
Multimodal Large Language Models (MLLMs) have drawn increasing attention for their powerful perceptual and reasoning capabilities \citep{Others:2024:OpenAI,ArXiv:2024:Wu:DeepSeekVL,ArXiv:2025:Lu,Arxiv:2025:Bai,Others:2025:Seed,Others:2025:Google}. These capabilities lay the foundation for developing autonomous agents that can interact with Graphical User Interfaces (GUIs) to accomplish complex real-world tasks \citep{ArXiv:2024:Zhang,ArXiv:2024:Nguyen,Arxiv:2025:Tang}. Early efforts in developing GUI agents primarily process structured interface descriptions (e.g.,~HTML), which significantly differs from human-computer interaction, and fails to generalize to novel scenarios without structured data \citep{ICLR:2023:Gur,NeurIPS:2023:Deng,KDD:2024:Lai}. For this reason, recent research has shifted toward \textit{native screen-based GUI agents} that operate directly on raw screen visuals \citep{Others:2024:Anthropic:CU,Others:2025:OpenAI,ArXiv:2025:Qin}.

Despite the growing progress, building advanced screen-based GUI agents is faced with the challenges of \textit{costly data curation}, \textit{sparse step-wise supervision}, and \textit{unstable optimization}. Specifically, first, empowering MLLMs with competent GUI navigation capabilities typically requires substantial amounts of high-quality interaction trajectories \citep{ArXiv:2024:Wang:EANT,ArXiv:2024:Wu:ATLAS,Arxiv:2025:Pahuja}. Different from conventional image-text data, these trajectories must capture not only the environmental states but also the detailed reasoning and corresponding actions. Therefore, constructing such high-quality trajectories relies heavily on expert involvement, leading to high costs in data organization and annotation. Second, completing a task in GUI environments typically demands multi-round interactions, resulting in long-horizon trajectories where feedback is only available at the final step. As a result, supervision information tends to be sparse and delayed, providing limited guidance for optimizing intermediate steps within the trajectory \citep{Arxiv:2025:Zhang:ProgRM,Arxiv:2025:Tang:G2}. Third, existing strategies for training GUI agents typically employ the Reinforcement Learning with Verifiable Rewards (RLVR) framework \citep{Arxiv:2025:Luo,Arxiv:2025:Wang}. However, due to the delayed nature of supervision information, the policy optimization must remain tightly coupled with the GUI agent execution environments, which are often non-stationary and thereby, lead to low training efficiency and unstable training process.

To handle the above challenges, we introduce a novel framework for native screen-based GUI agents that enables manageable data curation, and supports stable and efficient optimization. Different from the commonly-used RLVR framework, our basic idea is to train an agentic-Q model, and employ it to evaluate each step-wise action within the trajectory. In this way, on the one hand, we alleviate the issue of sparse supervision in GUI-agent scenarios; on the other hand, we decouple policy optimization from the environment, thereby mitigating the impact of environmental non-stationarity on the training process. Our framework incorporates two key components: 
\begin{itemize} 

    \item \textbf{Agentic-Q Estimation.} Our agentic-Q model is employed to evaluate an individual action given the associated state in GUI environments. To this end, we begin by collecting a set of self-generated state–action trajectories, and propagate the final supervision back to each intermediate step along the trajectory. These step-level annotations are then used to train our agentic-Q model in a binary classification manner. To ensure that the Q-model generalizes well, we incorporate \textit{sliding-window} and \textit{action-focus} strategies during training, which are empirically verified to mitigate potential entropy collapse when the agentic-Q model is subsequently used to guide policy optimization;

    \item \textbf{Step-Wise Policy Optimization.} 
    With the agentic-Q model, we can conveniently leverage step-wise trajectories generated by the policy itself for fine-grained policy optimization. Specifically, at each step, the agentic-Q model evaluates the action taken by the current policy under the corresponding state, which is then employed to refine the policy via reinforcement learning. Algorithmically, we adopt step-wise variants of critic-free methods, i.e.,~GRPO \citep{ArXiv:2024:Shao}, RLOO \citep{ArXiv:2024:Ahmadian}, and REINFORCE++ \citep{ArXiv:2025:Hu}, which are integrated with \textit{action-level return-level filtering}, and \textit{adaptive group weighting}, to ensure stable policy optimization.

\end{itemize}

We would like to highlight that the feasibility of our agentic-Q estimation is grounded in the natural interaction patterns of GUI agents, which differ substantially from other reasoning-decision tasks, e.g.,~mathematical solving and code generation. To be specific, GUI agents operate in multi-turn interactive settings, where both state transitions and actions are explicitly defined and observable. Each state directly reflects the visual feedback of the environment (i.e.,~the webpage layout), so that the observation space is structurally organized and semantically informative. Moreover, the action space is finite and task-specific, typically limited to a small set of atomic operations such as \textit{clicking}, \textit{scrolling} and \textit{typing}, which inherently supports tractable agentic-Q estimation. Notably, our framework enjoys two favorable advantages: (i) all state-action trajectories used for training are self-generated by the policy through interactions with the GUI environment, without the need for costly expert annotations \nocite{NeurIPS:2025:Wang:SPACE,NeurIPS:2025:Wang:TSPIN}; (ii) since the agentic-Q model offers intermediate supervision at each decision step, policy updates do not rely on delayed trajectory-level feedback. As a result, policy optimization is fully decoupled from non-stationary execution environment, thereby avoiding the instability and inefficiency.

Empirically, to evaluate performance in realistic scenarios, we build a real-time environment where policies can interact with live websites, and employ WebVoyager \citep{ACL:2024:He} and Online-Mind2Web \citep{Arxiv:2025:Xue} as benchmarks, both of which comprise tasks constructed from real-world websites. Additionally, we also include ScreenSpot \citep{ACL:2024:Cheng:SeeClick} to assess the fundamental grounding capability of GUI agents. We employ Ovis2.5-9B \citep{ArXiv:2025:Lu} as the base model, and use it to generate state–action trajectories through interactions with websites from WebVoyager. Overall, experimental results in Figure~\ref{fig:benchmark} demonstrate that our framework delivers remarkable performances, achieving state-of-the-art results among models of similar size (e.g.,~Qwen3-VL-8B and UI-TARS 1.5-7B) and also remaining competitive with significantly larger frontier models (e.g.,~GPT-4o and Claude 3.7/4 Sonnet CU). Additionally, since the training data only includes websites from WebVoyager, the results on Online-Mind2Web actually reflect out-of-distribution performances. As shown in Figure~\ref{fig:benchmark}, our models also achieve superior performances on Online-Mind2Web, highlighting their powerful generalization capability.
\section{Preliminaries} \label{sec:preliminary}
\textbf{Formulation.} 
The GUI navigation can be formulated as a Markov Decision Process (MDP): 
$$
    \M = (\S, \A, \P, \R)
$$ 
with a state space $\S$, an action space $\A$, a transition function $\P$ and a reward oracle $\R$. Here, we consider the undiscounted setting with a discount factor $\gamma=1$, and thereby omit $\gamma$ from the formulation for brevity. Specifically, at each round $i$, the agent $\pi_\theta$ parameterized by $\theta$ receives a state $s_i$ from the space $\S$, and chooses an action $a_i \in \A$ based on its own thought $t_i$. After that, the environment updates the state to 
$
    s_{i+1}\sim \P(s_{i+1}|s_i, t_i, a_i)
$
so that the agent $\pi_\theta$ can select the action $a_{i+1}$ for the next round. This process is repeated until the agent $\pi_\theta$ considers the task complete or reaches a terminal state. At the end, the agent $\pi_\theta$ receives a reward $r$ from $\R$ that indicates whether the task successfully finishes.

In this framework, (i) the state $s_i \in \S$ includes a task query, user instructions, historical interactions and the screenshot at round $i$; (ii) the action $a_i$ is chosen from a discrete and finite space $\A$ that spans common web-use operations, such as \textit{left-click} and \textit{scroll}. Detailed descriptions can be found in Appendix~\ref{sec:prompts}; (iii) the thought $t_i$ is a text produced by $\pi_\theta$ that summarizes the reasoning and planning for action $a_i$. We define the \textit{step} of the iterative process as a tuple $(s_i, t_i, a_i, r_i)$ at the round $i$. Then, the whole trajectory with total rounds of $T$ can be represented as: 
$$
    \zeta = \{(s_1, t_1, a_1, r_1), (s_2, t_2, a_2, r_2), \dots, (s_T, t_T, a_T, r_T)\}.
$$
Note that only a terminal reward $r_T$ is obtained (indicating whether the task is successfully finished), and step-wise rewards $r_1, \dots, r_{T-1}$ are unavailable. To avoid introducing additional assumptions about the unobserved rewards, we set $r_i = 0~(i < T)$ for all intermediate steps. Under this setting, the return at step $i$ reduces to $G_i = \sum_{j=i}^T{\gamma^{j-i} r_j} = r_T$ with the choice of $\gamma=1$. Consequently, its expectation $\E[G_i] = P(r_T = 1 \mid s_i,t_i,a_i, \pi_\theta)$ measures the probability for the task success given the first $i$ steps.

\textbf{Proximal Policy Optimization (PPO).} 
Reinforcement Learning (RL) serves as a fundamental paradigm for improving agent–environment interactions \citep{NeurIPS:2022:Ouyang}. A widely-used algorithm in RL for large language models is PPO \citep{ArXiv:2017:Schulman}, which employs an actor-critic framework to update the policy iteratively. Formally, PPO aims to maximize the following objective: 
\begin{equation*}
    \J_{\ppo}(\theta) = \E_{\x \sim q(\cdot), \y \sim \pi_{\theta_\old}(\cdot|\x)} \left[ \frac{1}{|\y|} \sum\nolimits_{t=1}^{|\y|} \min(w_t(\theta) A_t, \clip(w_t(\theta), 1-\epsilon, 1+\epsilon) A_t) \right],
\end{equation*}
where $\pi_\theta$, $\pi_{\theta_\old}$ denote the current and old policies, and $\x \sim q(\cdot)$, $\y \sim \pi_{\theta_\old}$ denote the prompt and the corresponding response. The importance sampling ratio $w_t(\theta) = \frac{\pi_\theta(y_t|\x, \y_{<t})}{\pi_{\theta_\old}(y_t|\x, \y_{<t})}$ is introduced for unbiased estimation, and the clipping hyper-parameter $\epsilon$ is employed for stable optimization. The advantage $A_t$ measures the relative improvement of the expected return for $y_t$ over a baseline that is estimated by a dedicated critic policy. Typically, the critic policy is of comparable scale to $\pi_\theta$, which brings substantial memory and computational overhead, particularly for large language model scenarios.

\textbf{Critic-Free Methods.} 
Recently, critic-free methods have emerged as efficient alternatives to PPO by eliminating the need for a separate critic policy. One well-known method is Group Relative Policy Optimization (GRPO) \citep{ArXiv:2024:Shao}, which computes the baseline in $A_t$ using the average reward across a group of responses generated for the same prompt. Mathematically, GRPO aims to optimize
\begin{equation*}
    \J_{\grpo}(\theta) = \E_{\x \sim q(\cdot), \{\y_i\}_{i=1}^K \sim \pi_{\theta_\old}(\cdot|\x)} \left[ \frac{1}{K} \sum\nolimits_{i=1}^{K} \frac{1}{|\y_i|} \sum\nolimits_{j=1}^{|\y_i|} \min(w_{i, j}(\theta) A_{i, j}, \clip(w_{i, j}(\theta), 1-\epsilon, 1+\epsilon) A_{i, j})\right]
\end{equation*}
where $K$ is the group size, and the advantage $A_{i, t}$ is computed by:
\begin{equation}
    \label{eq:grpo:advantage}
    A_{i, t} = A_{i} = \frac{r_i - \mean(r_1, \dots, r_K)}{\std(r_1, \dots, r_K)}
\end{equation}
with the reward $r_i$ for the $i$-th response in the group. Following the similar idea, REINFORCE Leave-One-Out (RLOO) \citep{ArXiv:2024:Ahmadian} computes the advantage by 
\begin{equation}
    \label{eq:rloo:advantage}
    A_{i, t} = A_{i} = r_i - \frac{1}{K-1} \sum\nolimits_{j\neq i} r_j = \frac{K}{K-1} \left(r_i - \mean(r_1, \dots, r_K)\right)
\end{equation}
to ensure unbiased estimation. REINFORCE++ \citep{ArXiv:2025:Hu} proposes to employ global batch statistics for normalization:
\begin{equation}
    \label{eq:rf++:advantage}
    A_{i, t} = \frac{A'_{i, t} - \mean_{\batch}(A')}{\std_{\batch}(A')},~\text{with}~A'_{i, t} = r_i - \mean(r_1, \dots, r_K).
\end{equation}
Notably, although these methods have proven effective in the RLVR framework, it is unsuitable to combine them with our agentic-Q directly. The reasons lie in that: (i) in the RLVR framework, rewards are typically binary, and these methods (e.g.,~GRPO) can naturally suppress extremely easy and hard tasks by vanished advantages. In our framework, we employ continuous rewards from our agentic-Q to optimize the agent. In extreme cases, small numerical differences in rewards can lead to low std, which in turn induce disproportionately large normalized advantages and result in unstable policy optimization; (ii) these methods optimize the objective by uniformly averaging over different groups, overlooking intrinsic differences of each group. Compared to easy tasks, hard ones are often more informative, but uniform weighting may ignore their contributions, leading to suboptimal optimization.
\section{Method}
\label{sec:q-learning}
\begin{figure}[t]
    \centering
    \includegraphics[width=1.0\textwidth]{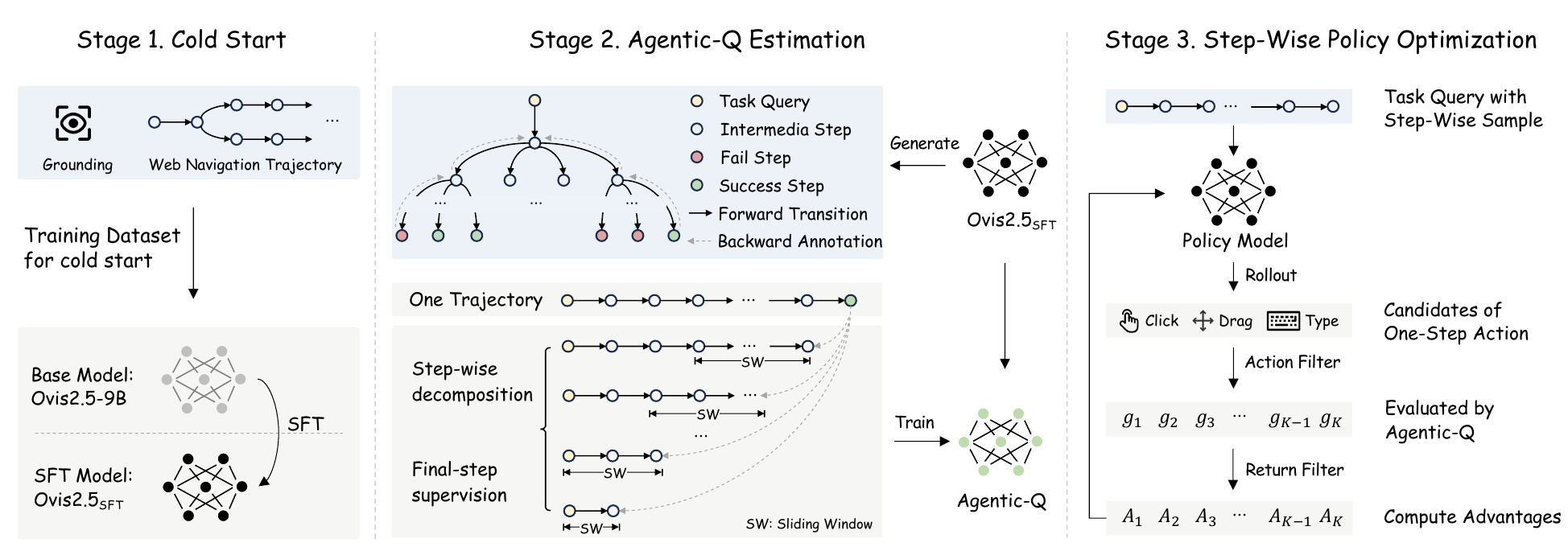}
    \caption{
        Illustration of our framework with three stages: (i) $\text{Ovis2.5}_\text{SFT}$ is trained on grounding data and a set of expert web navigation trajectories; (ii) we collect state-action trajectories by $\text{Ovis2.5}_\text{SFT}$ itself, and train our agentic-Q model in a binary classification manner; (iii) we optimize the policy with self-generated step-wise trajectories under the guidance of our agentic-Q model.
    }
    \label{fig:pipline}
\end{figure}

This section presents our framework for training a GUI agent, with the overall pipeline illustrated in Figure~\ref{fig:pipline}. In the subsequent parts, we begin by describing the details of data curation and cold-start training, followed by agentic-Q estimation and step-wise policy optimization.

\subsection{Data Curation}
\textbf{Detection-Rerunning Strategy.} 
The environment where GUI agents operate is typically non-stationary, posing significant challenges for data collection and performance evaluation. For example, due to load limitations and system configurations, it is common to encounter environmental issues including human verification, inaccessible websites and browser crashes when constructing web-browsing trajectories. In these cases, task-irrelevant noise is inadvertently introduced when collecting training data, which hinders the model from capturing the underlying task-specific patterns. Moreover, failed tasks in these cases cannot be directly attributed to insufficient model performance during evaluation. To address these issues, we use a simple yet effective strategy, namely detection-rerunning. Specifically, we employ expert models to assess whether a given trajectory has been affected by environmental issues. If such disturbances are detected, we rerun the corresponding task, and reconstruct the trajectory accordingly. In practice, we find that this simple strategy can mitigate almost all environmental issues.

\noindent
\textbf{Data Stratification.}
To ensure a stable and effective optimization on agent models, we conduct data stratification over the trajectories in training set. Specifically, for a single task, we repeat the web-browsing $n = 8$ times, and rank the task into Level $l \in \{0, 1, \dots, 8\}$ based on the number of successful trajectories, where Level $l$ indicates that $l$ out of $n = 8$ rollouts successfully handle the task. It is important to note that tasks at Level $0$ (highly challenging) and Level $8$ (trivially easy) yield the same results (either all failures or all successes) across repeated rollouts, which contribute minimally to enhancing exploration. Therefore, we exclude the two types of tasks from the training set.

\noindent
\textbf{Progress Supervision Annotation.}
In state-action trajectories, only the final-step supervision on the last action is available, resulting in a lack of process-level guidance for intermediate actions and posing challenges for training the agentic-Q model. A natural attempt is to annotate intermediate supervision using advanced models. However, this approach incurs substantial annotation costs, and may fail to reflect the underlying contribution of each step to the final result. To address the scarcity of process supervision, we propose to propagate the final outcome backward to each intermediate step, treating it as a form of process-level label for training agentic-Q model. Formally, for a trajectory $\zeta = \{(s_i, t_i, a_i, r_i)\}_{i=1}^T$, the return $G_{i} = r_T$ corresponding to the first $i$ steps serves as an unbiased estimator of the probability for task completion. It is therefore natural to approximate $G_i$ in an empirical-risk-minimization manner.
\subsection{Cold Start}
To initialize the model $\pi_\theta$ with fundamental GUI navigation capabilities, we fine-tune $\pi_\theta$ on grounding data and web navigation trajectories from experts, in which the trajectory data only records the browsing behavior without supervision on task completion. We adopt supervised fine-tuning (SFT) for cold start. For example, given web navigation trajectory data $\zeta = \{(s_i, t_i, a_i)\}_{i=1}^T$, the goal of SFT is to minimize:
\begin{equation}
    \label{eq:sft}
    \L_{\sft} = - \E_{\{(s_i, a_i, t_i)\}\sim{\D_\sft}} \left[ \sum\nolimits_{i=1}^{T} \log \pi_\theta (t_i, a_i|s_i) \right], \nonumber
\end{equation}
where $\D_\sft$ is the trajectory dataset for SFT. After the cold start, our MLLM-based agent acquires basic GUI navigation capabilities, such as understanding real-world webpages and generating valid actions (e.g.,~\textit{left-click} and~\textit{scroll}). Starting from this initial model $\pi_{\theta_\sft}$, we proceed to train an agentic-Q model, and further enhance GUI navigation abilities via reinforcement learning.
\subsection{Agentic-Q Estimation}
The agentic-Q estimation begins by collecting a set of self-generated trajectories under $\pi_{\theta_\sft}$, with the process supervision propagated from the final outcome. Then, instead of treating the whole trajectory as a single sample, we decompose it into step-wise segments. Specifically, we employ each step $(s_i, t_i, a_i, r_i)$ in the trajectory $\zeta = \{(s_i, t_i, a_i, r_i)\}_{i=1}^T$ as an independent training instance. The goal of our agentic-Q model $Q_\theta$ is to predict the return $G_i$ corresponding to the step $(s_i, t_i, a_i)$. Since $G_i = r_T \in \{0, 1\}$ is binary, a natural choice for agentic-Q estimation is to minimize the cross-entropy loss:
\begin{equation}
    \L_{\ce} = - \E_{(s_i, t_i, a_i, G_i)\sim\D_{\step}} \left[ G_i \log Q_{\theta} (s_i, t_i, a_i) + (1 - G_i) \log (1 - Q_{\theta} (s_i, t_i, a_i)) \right], \nonumber
\end{equation}
where the dataset $\D_{\step}$ contains step-wise samples. According to the favorable property of cross-entropy loss, the agentic-Q model with optimal parameters can provably capture the distribution of return.

To ensure the agentic-Q model generalizes well and to support stable policy optimization, we incorporate two techniques: \textit{sliding-window} and \textit{action-focus}. Specifically, at each step $i$, the state $s_i$ encodes all previous interactions. When the interaction history becomes too long, the agentic-Q model struggles to focus on the current context, leading to instability in subsequent policy optimization. For this reason, we adopt a sliding-window strategy that restricts the historical interactions in $s_i$ to only the most recent ones. Moreover, since the step-wise sample includes both the thought and the action at each step, we apply an action-focus strategy that masks out the thought part during training. This encourages the agentic-Q model to focus on the action itself, reducing potential disturbance from thought part.
\subsection{Step-Wise Policy Optimization (SWPO)}
Based on the initial model $\pi_{\theta_\sft}$ and self-generated step-wise trajectories $\D_{\step}$, we proceed to further enhance the GUI navigation capability of our model. To this end, we propose a critic-free step-wise policy optimization method, combined with the agentic-Q model $Q_\theta$. Compared to existing policy optimization methods, such as GRPO \citep{ArXiv:2024:Shao}, we incorporate the following modifications:
\begin{itemize}
    \item Instead of using final-outcomes as supervisions, we employ the predictions of $Q_\theta$ for policy optimization. In this way, we can conveniently use step-wise samples, and decouple the GUI execution environment from the policy optimization, thereby improving the efficiency;

    \item We introduce an action-level return-level filtering strategy. To be specific, given a state $s$, the agent model $\pi_\theta$ rolls out $K$ candidate actions, and the agentic-Q model predicts the return for each action. If all proposed actions are identical (i.e.,~lack diversity), we discard the sample $s$; similarly, if the $K$ returns are highly concentrated (i.e.,~with low std), we also filter out $s$;

    \item We explicitly consider the task difficulty by introducing adaptive group weighting. Inspired by self-normalized inverse propensity scoring \citep{NeurIPS:2015:Swaminathan}, we use the inverse 1259of average return for a group as the weight, to draw more attention to harder tasks.

\end{itemize}
The above designs can naturally integrate with existing critic-free policy optimization methods. Taking GRPO as an example, the objective of the step-wise variant of GRPO can be formulated as:
\begin{equation*}
    \J_{\sgrpo}(\theta) = \E\left[ \frac{u(s)}{K} \sum\nolimits_{i=1}^{K} \frac{1}{|t_i| + |a_i|} \sum\nolimits_{j=1}^{|t_i| + |a_i|} \min(w_{i, j}(\theta) A_{i, j}, \clip(w_{i, j}(\theta), 1-\epsilon, 1+\epsilon) A_{i, j}) \right]
\end{equation*}
where the state $s \sim q'(\cdot)$ is sampled from the filtered set $q'(\cdot)$, and the expectation is taken over $s \sim q'(\cdot)$ and $\{(t_i, a_i)\}_{i=1}^K \sim \pi_{\theta_\old}(\cdot|s)$. 
The advantage $A_{i, j}$ and the group weight $u(s)$ are computed by:
\begin{equation*}
    A_{i, j} = A_{i} = \frac{g_i - \mean(g_1, \dots, g_K)}{\std(g_1, \dots, g_K)}~\text{and}~u(s) = \frac{\bar{g}(s)^{-1}}{\sum_{s'} \bar{g}(s')^{-1}},
\end{equation*}
where $g_i$ denotes the return from $Q_\theta$ for the step $(s, t_i, a_i)$ with the $i$-th reasoning $t_i$ and action $a_i$, and $\bar{g}(s)$ denotes the average return for $s$. Analogously, we can also derive step-wise variants of RLOO \citep{ArXiv:2024:Ahmadian} and REINFORCE++ \citep{ArXiv:2025:Hu}, denoted as $\srloo$ and $\srf$, respectively.
\section{Experiments}
In this section, we present empirical studies to validate the effectiveness of our framework. We commence by describing the experimental setup, and then deliver the main results with corresponding analyses.

\subsection{Experimental Setup}
Our framework is built on Ovis2.5-9B \citep{ArXiv:2025:Lu} which comprises a 970M-parameter visual encoder (i.e.,~a native-resolution ViT and a carefully-designed VET) and a 8B-parameter LLM (i.e.,~Qwen3-8B). 

\noindent
\textbf{Benchmark.} We conduct evaluations on a diverse set of benchmarks that cover two categories: (i) For GUI grounding, ScreenSpot \citep{ACL:2024:Cheng:SeeClick} provides over $1200$ instructions and $600$ screenshots, covering both text-based elements and a variety of widgets and icons across mobile, desktop and webpage; (ii) For GUI navigation, WebVoyager \citep{ACL:2024:He} compiles tasks from real-world websites (e.g., Amazon and Apple), and employs GPT-4V \citep{ArXiv:2023:OpenAI} as an automated evaluator to assess the performance of GUI agents. Online-Mind2Web \citep{Arxiv:2025:Xue} offers more diverse tasks of varying difficulty levels (easy, medium, and hard), collected from $136$ real-world websites. Both benchmarks are evaluated in real-time environments where GUI agents interact with live websites, measuring authentic navigation capabilities.

\noindent
\textbf{Contenders.} We compare our models against a set of advanced models with varying sizes, such as 
GPT-4o \citep{Others:2024:OpenAI}, Claude 3.7/4 Sonnet Compute Use \citep{Others:2025:Anthropic:Claude3.7,Others:2025:Anthropic:Claude4}, Qwen3-VL-8B/32B \citep{Arxiv:2025:Bai}, UI-TARS-7B/72B \citep{ArXiv:2025:Qin} and UI-TARS 1.5 7B \citep{Others:2025:Seed:UI-TARS}.

\subsection{Experimental Results}
We present the experimental results for GUI grounding in Table~\ref{table:ss}, and for GUI navigation in Tables~\ref{table:wv}~and~\ref{table:omw}.

\begin{table}[t]
    \centering
    \small
    \renewcommand{\arraystretch}{1.4}
    \setlength{\tabcolsep}{5pt} 
    \setlength{\aboverulesep}{0pt}
    \setlength{\belowrulesep}{0pt}
    \caption{
        Performance (\%) comparisons on ScreenSpot v1 and v2, where $\text{Ovis2.5}_\text{S-GRPO}$, $\text{Ovis2.5}_\text{S-RLOO}$ and $\text{Ovis2.5}_\text{S-RF++}$ are trained by step-wise variants of GRPO, RLOO and REINFORCE++, respectively. At the last column of each version, we report the average accuracy (Avg) over all scenarios. 
    }
    \label{table:ss}
    \resizebox{1.0 \textwidth}{!}{
    \begin{tabular}{l cccccc >{\columncolor{tableblue}}c cccccc >{\columncolor{tableblue}}c} 
        \toprule
        & \multicolumn{7}{c}{\textbf{ScreenSpot v1}} & \multicolumn{7}{c}{\textbf{ScreenSpot v2}} \\ 
        \cmidrule(lr){2-8} \cmidrule(lr){9-15}

        & \multicolumn{2}{c}{Mobile} & \multicolumn{2}{c}{Desktop} & \multicolumn{2}{c}{Web} & 
        & \multicolumn{2}{c}{Mobile} & \multicolumn{2}{c}{Desktop} & \multicolumn{2}{c}{Web} & \\ 
        \cmidrule(lr){2-3} \cmidrule(lr){4-5} \cmidrule(lr){6-7} \cmidrule(lr){9-10} \cmidrule(lr){11-12} \cmidrule(lr){13-14}

        \multirow{-3}{*}{\textbf{Model}} & Text & I/W & Text & I/W & Text & I/W & \multirow{-2}{*}{\textbf{Avg}} 
        & Text & I/W & Text & I/W & Text & I/W & \multirow{-2}{*}{\textbf{Avg}} \\ 

        \midrule
        GPT-4o & $20.20$ & $24.90$ & $21.10$ & $23.60$ & $12.20$ & $7.80$ & $18.30$ & -- & -- & -- & -- & -- & -- & $20.10$ \\ 
        Claude 4 Sonnet CU & $97.07$ & $79.48$ & $81.44$ & $64.29$ & $92.61$ & $82.52$ & $84.75$ & $97.59$ & $82.94$ & $84.02$ & $67.14$ & $96.58$ & $82.27$ & $87.11$  \\ 
        Claude 3.7 Sonnet CU & $88.28$ & $73.36$ & $79.90$ & $62.86$ & $90.87$ & $80.58$ & $80.74$ & $89.31$ & $75.83$ & $82.47$ & $72.86$ & $93.59$ & $80.30$ & $83.57$  \\ 
        Qwen3-VL-8B &$95.24$&	$79.91$&	$94.85$&	$81.43$&	$92.17$	&$83.50$&	$88.44$& $97.93$&	$86.73$&	$95.88$&	$81.43$	&$94.02$&	$86.21$ & $91.35$   \\ 
        Qwen3-VL-32B &$96.34$	&$85.59$	&$96.91$	&$87.86$	&$90.87$&	$90.78$	&$91.67$& $98.28$&	$89.10$	&$98.97$&	$90.71$&	$94.87$&	$92.61$	&$94.50$  \\
        UI-TARS-7B & $92.31$&	$85.15$&	$90.72$	&$84.29$	&$88.70$	&$85.92$&	$88.21$& $94.83$	&$88.15$	&$93.81$&	$86.43$&	$93.16$	&$86.21$&	$90.96$   \\
        UI-TARS-72B & $94.14$&	$82.53$&	$95.36$&	$85.00$	&$90.87$&	$83.50$	&$88.92$& $96.90$&	$86.73$&	$97.42$	&$88.57$	&$91.88$&	$86.21$&	$91.75$  \\
        UI-TARS 1.5-7B & $92.67$ & $77.73$  & $93.81$  & $71.43$ & $86.52$  &  $77.18$ & $84.20$  & $96.55$ & $82.94$ & $96.91$ & $76.43$ & $90.17$ & $76.85$ & $87.81$ \\

        \midrule
        $\textbf{Ovis2.5}_\textbf{SFT}$ & $95.24$ & $78.60$ & $94.85$ & $85.00$ & $92.61$ & $83.01$ & $88.60$ & $97.24$ & $84.83$ & $97.42$ & $87.86$ & $95.73$ & $86.70$ & $92.22$ \\
        $\textbf{Ovis2.5}_\textbf{S-GRPO}$ & $94.51$ & $79.04$ & $94.85$ & $85.00$ & $92.61$ & $83.01$ & $88.52$ & $97.24$ & $85.31$ & $97.94$ & $87.86$ & $96.15$ & $86.21$ & $92.37$ \\
        $\textbf{Ovis2.5}_\textbf{S-RLOO}$ & $94.87$ & $79.04$ & $94.85$ & $85.00$ & $92.17$ & $82.52$ & $88.44$ & $97.24$ & $85.31$ & $97.42$ & $88.57$ & $95.73$ & $86.70$ & $92.37$ \\
        $\textbf{Ovis2.5}_{\textbf{S-RF++}}$ & $95.24$ & $79.04$ & $94.85$ & $84.29$ & $92.61$ & $83.50$ & $88.68$ & $97.59$ & $85.31$ & $97.42$ & $87.86$ & $95.73$ & $87.19$ & $92.45$ \\
        \bottomrule
    \end{tabular}
    }
\end{table}

\begin{table}[t]
    \centering
    \caption{
        Performance (\%) comparisons on WebVoyager, where three models $\text{Ovis2.5}_\text{S-GRPO}$, $\text{Ovis2.5}_\text{S-RLOO}$ and $\text{Ovis2.5}_\text{S-RF++}$ are trained by the step-wise variants of GRPO, RLOO and REINFORCE++, respectively. At the last column, we report the average task success rate over all websites, as well as the \textcolor{green}{improvements} of our three models over their SFT version $\text{Ovis2.5}_\text{SFT}$.
    }
    \small
    \renewcommand{\arraystretch}{1.5}
    \setlength{\tabcolsep}{2pt}
    
    \setlength{\aboverulesep}{0pt}
    \setlength{\belowrulesep}{0pt}
    \label{table:wv}
    \resizebox{\textwidth}{!}{
    \begin{tabular}{l cccccccccccc >{\columncolor{tableblue}}c} 
        \toprule
        \rule{0pt}{4ex} \textbf{Model} & \makecell{\textbf{Allrecipes}} & \textbf{Amazon} & \textbf{Apple} & \textbf{ArXiv} & \makecell{\textbf{BBC}\\\textbf{News}} & \makecell{\textbf{Cambridge}\\\textbf{Dictionary}} & \textbf{Coursera} & \textbf{ESPN} & \textbf{GitHub} & \makecell{\textbf{Google}\\\textbf{Map}} & \makecell{\textbf{Hugging-}\\\textbf{-face}} & \makecell{\textbf{Wolfram}\\\textbf{Alpha}} & \textbf{Avg} \\
        \midrule
        GPT-4o           & $55.56$ & $53.66$ & $55.81$ & $60.47$ & $54.76$ & $81.40$ & $64.29$ & $43.18$ & $58.54$ & $56.10$ & $41.86$ & $65.22$ & $57.59$ \\ 
        Claude 4 sonnet CU & $84.44$ & $75.61$ & $69.77$ & $81.40$ & $85.71$ & $95.35$ & $85.71$ & $86.36$ & $87.80$ & $78.05$ & $69.77$ & $84.78$ & $82.10$ \\
        Claude 3.7 sonnet CU & $75.56$ & $75.61$ & $62.79$ & $86.05$ & $69.05$ & $88.37$ & $78.57$ & $75.00$ & $68.29$ & $85.37$ & $72.09$ & $82.61$ & $76.65$ \\
        Qwen3-VL-8B      & $35.56$ & $46.34$ & $34.88$ & $34.88$ & $40.48$ & $67.44$ & $61.90$ & $27.27$ & $48.78$ & $60.98$ & $32.56$ & $50.00$ & $44.94$ \\
        Qwen3-VL-32B     & $37.78$ & $26.83$ & $46.51$ & $48.84$ & $54.76$ & $79.07$ & $64.29$ & $54.55$ & $65.85$ & $56.10$ & $18.60$ & $34.78$ & $48.83$ \\
        UI-TARS-7B       & $17.78$ & $31.71$ & $16.28$ & $20.93$ & $40.48$ & $58.14$ & $71.43$ & $45.45$ & $31.71$ & $17.07$ & $41.86$ & $39.13$ & $35.99$ \\
        UI-TARS-72B      & $15.56$ & $39.02$ & $46.51$ & $65.12$ & $40.48$ & $72.09$ & $59.52$ & $47.73$ & $43.90$ & $26.83$ & $25.58$ & $47.83$ & $44.16$ \\
        UI-TARS 1.5-7B   & $26.67$ & $29.27$ & $48.84$ & $53.49$ & $61.90$ & $60.47$ & $66.67$ & $38.64$ & $41.46$ & $58.54$ & $46.51$ & $50.00$ & $48.44$ \\
        \midrule
        $\textbf{Ovis2.5}_\textbf{SFT}$        & $66.67$ & $70.73$ & $58.14$ & $79.07$ & $78.57$ & $88.37$ & $69.05$ & $56.82$ & $70.73$ & $60.98$ & $44.19$ & $80.43$ & $68.68$ \\
        $\textbf{Ovis2.5}_\textbf{S-GRPO}$       & $82.22$ & $78.05$ & $62.79$ & $76.74$ & $80.95$ & $90.70$ & $95.24$ & $65.91$ & $90.24$ & $87.80$ & $60.47$ & $76.09$ & $78.79_{\textcolor{green}{(+10.11)}}$ \\
        $\textbf{Ovis2.5}_\textbf{S-RLOO}$       & $88.89$ & $85.37$ & $67.44$ & $72.09$ & $76.19$ & $93.02$ & $83.33$ & $70.45$ & $82.93$ & $70.73$ & $62.79$ & $80.43$ & $77.82_{\textcolor{green}{(+9.14)}}$ \\
        $\textbf{Ovis2.5}_\textbf{S-RF++}$       & $77.78$  & $70.73$  & $69.77$  & $83.72$  & $80.95$  & $81.40$  & $92.86$  & $72.73$  & $75.61$  & $80.49$  & $72.09$  & $80.43$  & $78.21_{\textcolor{green}{(+9.53)}}$  \\
        \bottomrule
    \end{tabular}
    }
\end{table}

\begin{table}[ht]
    \centering
    \caption{
        Performance (\%) comparisons on Online-Mind2Web, where three models $\text{Ovis2.5}_\text{S-GRPO}$, $\text{Ovis2.5}_\text{S-RLOO}$ and $\text{Ovis2.5}_\text{S-RF++}$ are trained by the step-wise variants of GRPO, RLOO and REINFORCE++, respectively. At the last row, we report the average task success rate over all difficulty levels, as well as the \textcolor{green}{improvements} of our three models over their SFT version $\text{Ovis2.5}_\text{SFT}$.
    }
    \small
    \renewcommand{\arraystretch}{1.7} 
    \setlength{\tabcolsep}{2pt}      
    \setlength{\aboverulesep}{0pt}
    \setlength{\belowrulesep}{0pt}
    \label{table:omw}
    \resizebox{\textwidth}{!}{
    \begin{tabular}{lcccccc|cccc} 
        \toprule
        \rule{0pt}{4ex} 
         & {GPT-4o} & {\makecell{Claude 4\\Sonnet CU}} & {\makecell{Claude 3.7\\Sonnet CU}} & {\makecell{Qwen3-VL\\-8B}} & {\makecell{Qwen3-VL\\-32B}} &   {\makecell{UI-TARS\\1.5 7B}} & $\textbf{Ovis2.5}_\textbf{SFT}$ & $\textbf{Ovis2.5}_\textbf{S-GRPO}$ &  $\textbf{Ovis2.5}_\textbf{S-RLOO}$ & $\textbf{Ovis2.5}_\textbf{S-RF++}$\\
        \midrule
        
        \textbf{Easy}   & --      & $71.60$ & $72.84$ & $34.94$ & $38.55$  & $57.83$ & $66.27$ & $77.11$ & $75.90$ & $78.31$ \\
        \textbf{Medium} & --      & $65.47$ & $66.91$ & $25.87$ & $27.97$  & $27.27$ & $51.75$ & $62.94$ & $60.84$ & $65.73$ \\
        \textbf{Hard}   & --      & $53.52$ & $46.48$ & $9.46$ & $14.86$  & $17.57$ & $39.19$ & $51.35$ & $51.35$ & $44.59$ \\
        
        \midrule
        \rowcolor{tableblue}
        \textbf{Avg}    & $37.00$   & $62.33$ & $61.00$ & $24.33$ & $27.67$ & $33.33$ & $52.67$ & $64.00_{\textcolor{green}{(+11.33)}}$ & $62.67_{\textcolor{green}{(+10.00)}}$ & $64.00_{\textcolor{green}{(+11.33)}}$ \\
        
        \bottomrule
    \end{tabular}
    }
\end{table}

\textbf{GUI Grounding.}
Overall, our models demonstrate remarkable performances compared to their powerful contenders across the scenarios of mobile, desktop and web, shown in Table~\ref{table:ss}. Specifically, the cold-start model $\text{Ovis2.5}_\text{SFT}$ achieves an average score of $88.60$ on ScreenSpot v1 and $92.22$ on ScreenSpot v2, outperforming similar-scale baselines (i.e.,~Qwen3-VL-8B, UI-TARS-7B and UI-TARS 1.5-7B), and even surpassing stronger proprietary models (i.e., GPT-4o, Claude 3.7/4 Sonnet). Moreover, we also observe that although step-wise policy optimization is performed on interaction trajectory data, the grounding ability of models does not degrade. In contrast, it even exhibits a slight improvement, such as $\text{Ovis2.5}_{\text{S-RF++}}$ achieves $92.45$ on ScreenSpot v2, compared to $92.22$ by $\text{Ovis2.5}_\text{SFT}$.

\textbf{GUI Navigation.}
Tables~\ref{table:wv}~and~\ref{table:omw} summarize performance comparisons on WebVoyager and Online-Mind2Web, respectively. From the results, we observe that our framework significantly improves GUI navigation performance across both benchmarks, achieving gains of $9.14$--$10.11$ points on WebVoyager and $10.00$--$11.33$ points on Online-Mind2Web over $\text{Ovis2.5}_\text{SFT}$. Notably, our three models outperform almost all baselines over two benchmarks. Compared to UI-TARS series and Qwen3-VL series, our models demonstrate stronger performances, even surpassing larger-scale counterparts. Compared to proprietary models such as GPT-4o and Claude Sonnet series, our models also exhibit superior overall performance. Furthermore, our framework exhibits high computational efficiency, and significantly improves generalization performances. Specifically, (i) the incorporation of agentic-Q decouples the RL post-training from the external environment, rendering policy optimization efficient; (ii) all state-action trajectories for training are generated by the policy itself, eliminating the need for expensive annotations; and (iii) although the training data consists of websites from WebVoyager, our models achieve notable performance gains on the more diverse Online-Mind2Web. This demonstrates that our framework empowers agents with stronger generalization capability.

\begin{figure}[t]
    \centering
    \includegraphics[width=1.0\textwidth]{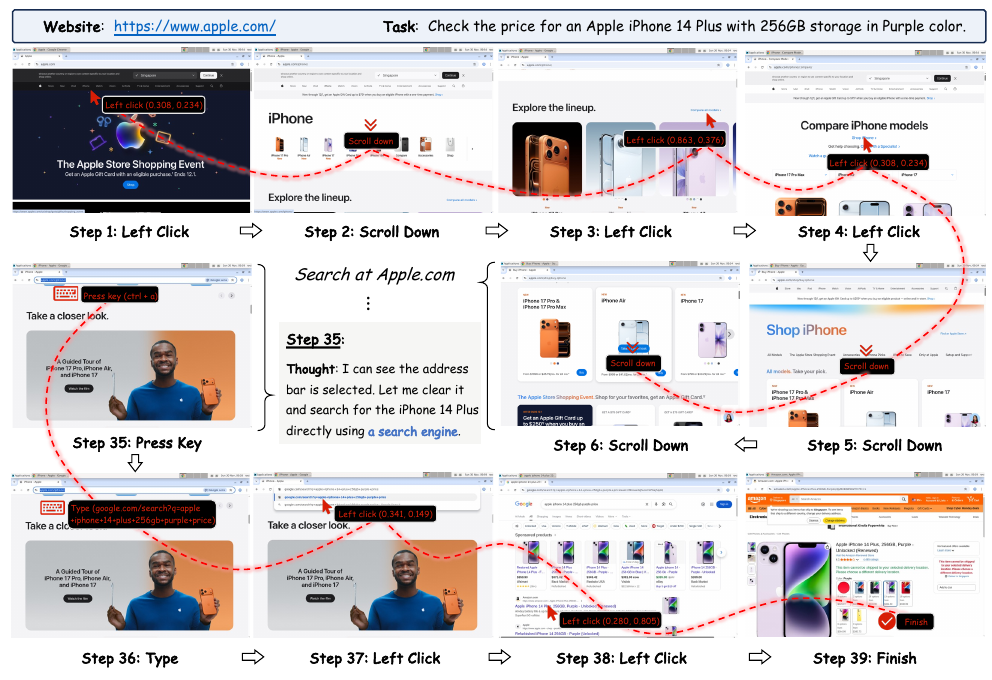}
    \caption{
        Example of boundary traversal. The GUI agent is prompted to retrieve the price of iPhone 14 Plus at \textit{www.apple.com}. However, due to recent updates, this product has been removed. For the first $34$ steps, the agent persistently searches within the site but fails to locate target information. At step $35$, it switches strategies and conducts an external search on \textit{Google}, completing the task by step $39$.
        }
    \label{fig:boundary_traversal}
\end{figure}

\begin{figure}[ht]
    \centering
    \includegraphics[width=1.0\textwidth]{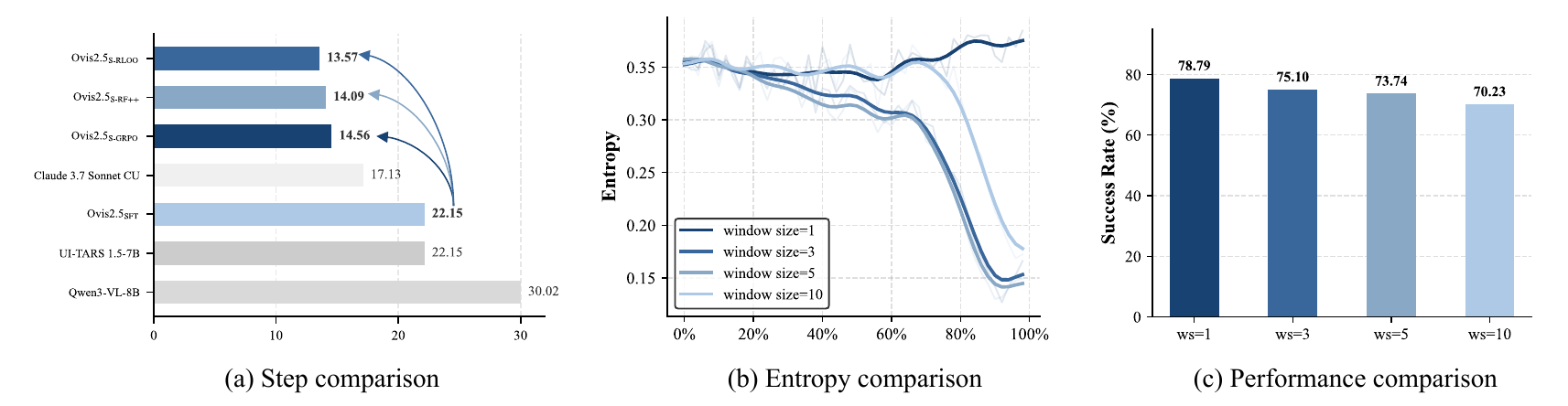}
    \caption{
        Comparisons under different setting: (a) compares average steps required to complete tasks in WebVoyager; (b) shows the policy entropy of $\text{Ovis2.5}_\text{S-GRPO}$ under varying sliding window sizes; (c) compares the overall performance of $\text{Ovis2.5}_\text{S-GRPO}$ across different window sizes.
    }
    \label{fig:step}
\end{figure}

\noindent
\textbf{Boundary Traversal.}
During the experiments, we observe an interesting behavior, termed as \textit{Boundary Traversal}: when the task is infeasible within the current website, our models are able to recognize the limitation and proactively switch to alternative solutions to complete the task successfully. We present a real-world example in Figure~\ref{fig:boundary_traversal}. Specifically, the agent is instructed to find the price of an iPhone 14 Plus on \texttt{www.apple.com}, but the product has been delisted due to website updates. According to the trajectory, the agent first attempts to locate the item by thoroughly searching within the Apple website. After $34$ unsuccessful steps, it realizes the task cannot be completed locally. At step $35$, the agent pivots to using a search engine, and by step $39$, successfully finds the desired information on Google. This behavior demonstrates a strong task-level generalization ability. Rather than being constrained by the boundaries of a single website, the agent can reason about task feasibility, identify failure conditions, and autonomously switch to alternative strategies. Furthermore, this behavior also exhibits the adaptive planning and goal-directed reasoning of the agent beyond local environments.

\noindent
\textbf{Navigation Efficiency.}
We evaluate the navigation efficiency of GUI agents by measuring the number of steps required for task completion, and present results on WebVoyager in Figure~\ref{fig:step}(a). The results show that our framework significantly improves navigation efficiency, e.g., reducing the number of steps from $22.15$ to $13.57$ for $\text{Ovis2.5}_{\text{S-RLOO}}$. We attribute this enhancement to the use of step-wise trajectories, which promotes deliberate action at each step and ensures efficient task completion \citep{NeurIPS:2024:Chen:Tool}.

\noindent
\textbf{Sliding Window.}
In experiments, we observe a strong positive correlation between the performance of GUI agent and the policy entropy, i.e., a sharp drop in entropy often leads to a significant decline in performances. Therefore, maintaining stable entropy is critical for policy optimization. To this end, when training and exploiting our agentic-Q, we propose the \textit{sliding window} technique, where returns are estimated based solely on the most recent steps. A key intuition is that using a sliding window limits the historical context, potentially mitigating premature convergence of the policy. This helps prevent early entropy collapse and reduces the risk of getting stuck in suboptimal solutions. Empirically, to verify the effectiveness of sliding window, we conduct ablation studies by varying the window size among $\{1, 3, 5, 10\}$ and training corresponding agentic-Q and policy models. The results are summarized in Figure~\ref{fig:step}(b) and (c). We observe that as the window size increases, the policy entropy becomes less stable and tends to decrease rapidly, which in turn leads to a noticeable degradation in performances. Therefore, in practice, we adopt $\textit{window size} = 1$ for more stable policy entropy and better performance.

\begin{table}[t]
    \centering
    \caption{
        Ablation studies of SWPO on WebVoyager, where $\text{GRPO}_\text{agentic-Q}$ is the baseline trained by GRPO with our agentic-Q, and the two variants are trained by sequentially adding \textit{action-level return-level filtering} and \textit{adaptive group weighting} to $\text{GRPO}_\text{agentic-Q}$. At the last column, we report the average task success rate over all websites, as well as the \textcolor{green}{improvements} over the baseline $\text{GRPO}_\text{agentic-Q}$.
    }
    \small
    \renewcommand{\arraystretch}{1.5}
    \setlength{\tabcolsep}{2pt}
    
    \setlength{\aboverulesep}{0pt}
    \setlength{\belowrulesep}{0pt}
    \label{table:ablation}
    \resizebox{\textwidth}{!}{
    \begin{tabular}{l cccccccccccc >{\columncolor{tableblue}}c} 
        \toprule
        \rule{0pt}{4ex} \textbf{Model} & \makecell{\textbf{Allrecipes}} & \textbf{Amazon} & \textbf{Apple} & \textbf{ArXiv} & \makecell{\textbf{BBC}\\\textbf{News}} & \makecell{\textbf{Cambridge}\\\textbf{Dictionary}} & \textbf{Coursera} & \textbf{ESPN} & \textbf{GitHub} & \makecell{\textbf{Google}\\\textbf{Map}} & \makecell{\textbf{Hugging-}\\\textbf{-face}} & \makecell{\textbf{Wolfram}\\\textbf{Alpha}} & \textbf{Avg} \\
        \midrule
        $\text{GRPO}_\text{agentic-Q}$    & $84.44$ & $65.85$ & $55.81$ & $79.07$ & $76.19$ & $81.40$ & $83.33$ & $59.09$ & $73.17$ & $68.29$ & $69.77$ & $82.61$ & $73.35$ \\	 	 	 	 	 
        + action-return filtering   & $80.00$ & $75.61$ & $62.79$ & $76.74$ & $78.57$ & $90.70$ & $88.10$ & $61.36$ & $80.49$ & $75.61$ & $72.09$ & $76.09$ & $76.46_{\textcolor{green}{(+3.11)}}$ \\
        + adaptive group weighting       & $82.22$ & $78.05$ & $62.79$ & $76.74$ & $80.95$ & $90.70$ & $95.24$ & $65.91$ & $90.24$ & $87.80$ & $60.47$ & $76.09$ & $78.79_{\textcolor{green}{(+5.44)}}$ \\
        \bottomrule
    \end{tabular}
    }
\end{table}

\noindent
\textbf{Ablation Studies on SWPO.}
We conduct ablation studies on SWPO to investigate the impact of its each component (i.e., \textit{action-level return-level filtering} and \textit{adaptive group weighting}). Specifically, starting from the baseline $\text{GRPO}_\text{agentic-Q}$ that is trained by GRPO with our agentic-Q, we construct two variants: one by introducing \textit{action-level return-level filtering}, and the other by further applying \textit{adaptive group weighting}. We report the results on WebVoyager in Table~\ref{table:ablation}. The average performance across different websites shows a $3.11$-point improvement when action-level return filtering is applied to the baseline, and $5.44$-point improvement when adaptive group weighting is further introduced. Notably, the gains are more significant on Coursera, GitHub, and Google Maps, highlighting the effectiveness of our two techniques.

\section{Conclusion}
\label{sec:conclusion}

In this report, we investigate native screen-based GUI agents, and propose a novel framework with two key designs: agentic-Q estimation and step-wise policy optimization. First, we train an agentic-Q model that can evaluate the contribution of a given action to task completion. Then, we optimize the agent model with our agentic-Q over step-wise state-action trajectories. All our training data are generated by the policy itself, so that the data collection costs are manageable. Moreover, the incorporation of agentic-Q decouples the policy update from the environment, ensuring stable and efficient optimization. Empirically, we have conducted comprehensive experiments to evaluate the performances on GUI grounding and GUI navigation. Experimental results demonstrate that our 9B agent achieves superior performance compared to others of the same scale, and even outperforms more powerful models such as the Claude series.

\bibliography{ref}
\bibliographystyle{plainnat}

\newpage
\appendix

\section{Trajectories on Websites}
In this section, we show trajectories of our agents on websites, e.g., Huggingface, Google Map, and Arxiv.

\begin{figure}[h]
    \centering
    \includegraphics[width=1.0\textwidth]{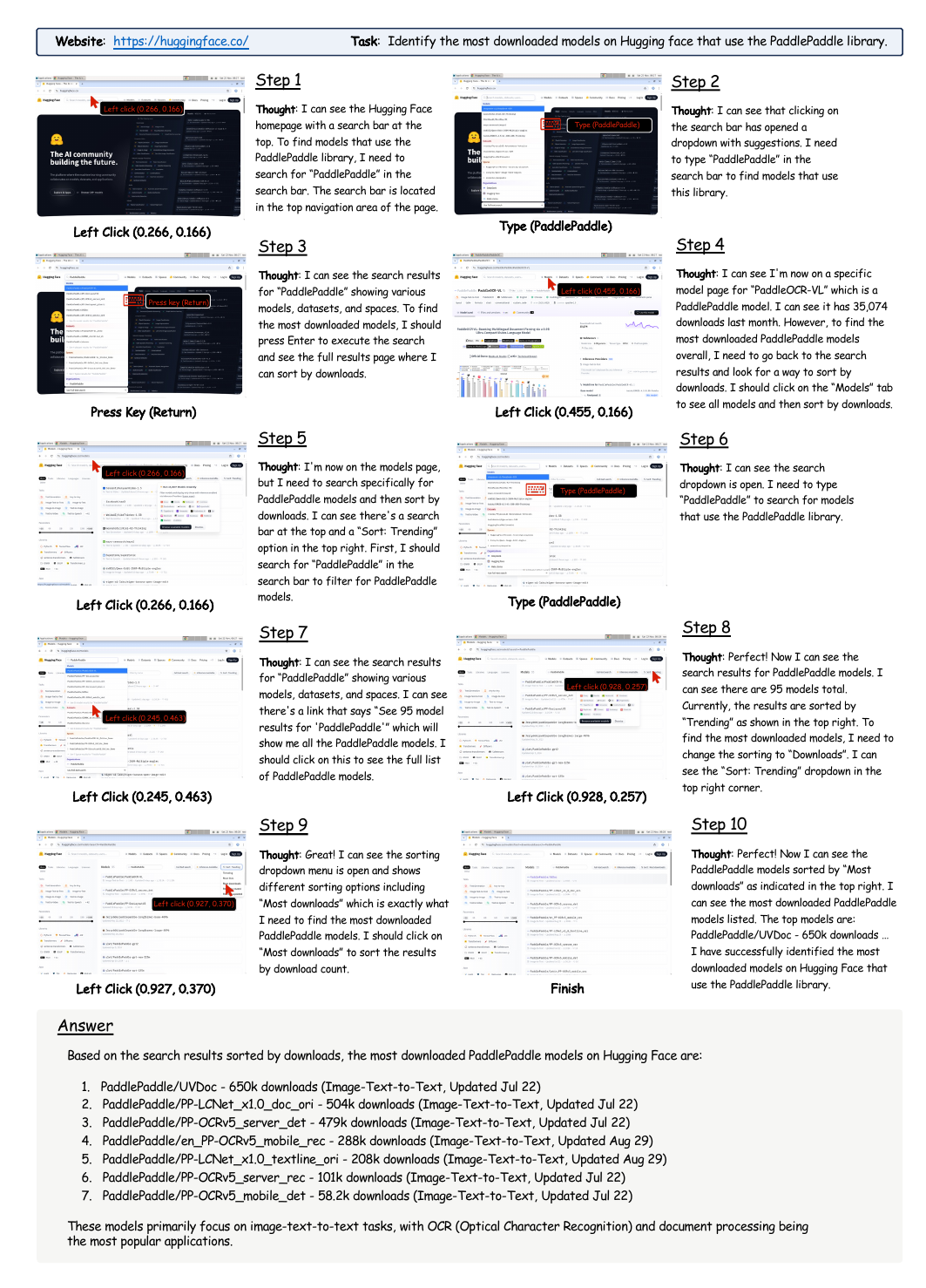}
    \caption{
        Trajectory for the task on Huggingface.
        }
    \label{fig:example-1}
\end{figure}

\newpage
\begin{figure}[h]
    \centering
    \includegraphics[width=1.0\textwidth]{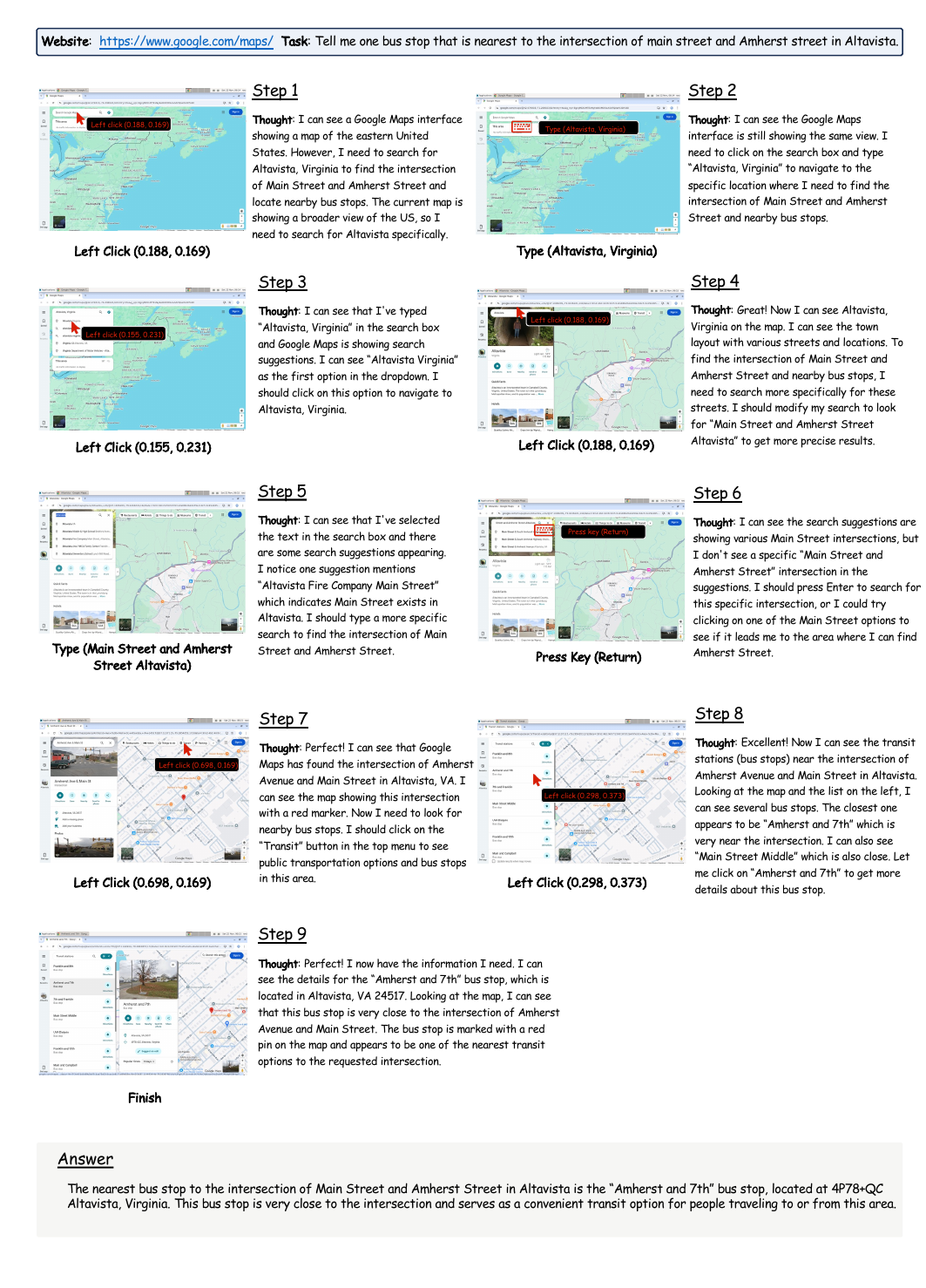}
    \caption{
        Trajectory for the task on Google Map.
        }
    \label{fig:example-2}
\end{figure}

\newpage
\begin{figure}[h]
    \centering
    \includegraphics[width=1.0\textwidth]{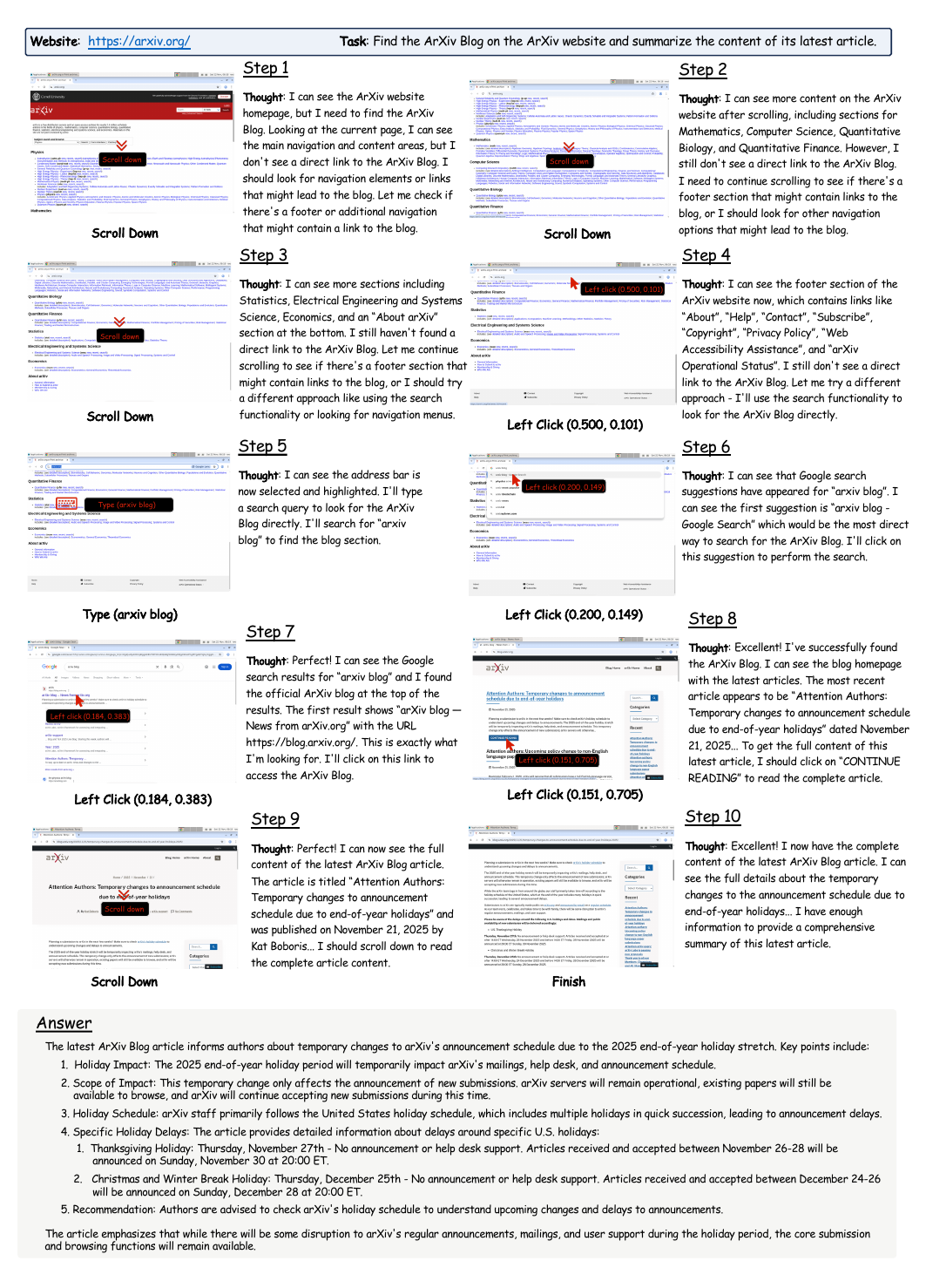}
    \caption{
        Trajectory for the task on Arxiv.
        }
    \label{fig:example-3}
\end{figure}

\newpage
\section{Prompts}
\label{sec:prompts}
In this section, we present the prompt (including descriptions of action space) used for our framework.
\begin{lstlisting}
    As a robot operating a browser inside a virtual machine, your task is to complete the query through a series of actions with the web page and strictly follow the rules.

      Core Requirements: 
      * Complete ALL requirements specified in the query. If multiple requirements exist, ensuring all are satisfied. 

      Rules:

      1. In each iteration, you will receive an Observation that includes a screenshot of the web page and some texts.

      2. First carefully analyze the visual information to determine the location corresponding to the web element that requires interaction. Describe the location using relative coordinates in the following format:
         <point>(x,y)</point>
         where x and y are floating-point numbers between 0.000 and 1.000 representing the horizontal and vertical positions, respectively.
         Note: If no specific point is required, output "None".

      3. Then choose one Action Type from the following actions:
         (1) HOVER: Hover the mouse over the specified coordinates.
         (2) LEFT_CLICK: Click the left mouse button on the specified coordinates.
         (3) RIGHT_CLICK: Click the right mouse button on the specified coordinates.
         (4) MIDDLE_CLICK: Click the middle mouse button on the specified coordinates.
         (5) DOUBLE_CLICK: Double-click the left mouse button on the specified coordinates.
         (6) TRIPLE_CLICK: Triple-click the left mouse button on the specified coordinates.
         (7) TYPE: Type the specified text.
         (8) DRAG: Drag the mouse from a start point to an end point. 
         (9) SCROLL: Scroll the page in the specified direction. If the coordinates are 'None', the default is to scroll the entire window. Otherwise, scroll from the specified coordinates.
         (10) WAIT: Wait for a specified duration or for unfinished page processes.
         (11) PRESS_KEY: Press a key or hotkey to complete a specific operation. The value supports xdotool's `key` syntax.
               Examples: "a", "Return", "alt+Tab", "ctrl+s", "Up", "KP_0" (for the numpad 0 key).
         (12) COPY_IMAGE: Get Image Url at the coordinates.
         (13) FINISHED: This action should be selected if the query has been resolved or is considered unresolvable.

      4. Finally, determine the action value based on the action type. Note that if no content is required, output 'None'.
         (1) HOVER: value is None.
         (2) LEFT_CLICK, (3) RIGHT_CLICK, (4) MIDDLE_CLICK, (5) DOUBLE_CLICK, (6) TRIPLE_CLICK, (12) COPY_IMAGE:
            For these click actions, the value is fixed to None. The specific coordinates to operate on are provided as the Action Element.
         (7) TYPE: value is the text content that needs to be entered.
         (8) DRAG: value is None. Action Element should be two <point>(x,y)</point> entries.
         (9) SCROLL: value is the scroll direction, such as left, right, up, or down.
         (10) WAIT: value is None.
         (11) PRESS_KEY: value is the key or key-combination to be pressed.
         (13) FINISHED: value is the final answer for the given query or the reason why it could not be solved.

      5. Output only one action per iteration and strictly in the following format:
         * Thought: {brief thoughts (brief summary of information that helps answer the query)}
         * Action Element: {the coordinates in <point>(x,y)</point> format; for DRAG operations, provide two coordinates; otherwise output 'None'}
         * Action Type: {one Action Type you choose to operate}
         * Action Value: {one Action Value you choose in the above format}

\end{lstlisting}

\end{document}